\pdfoutput=1 
\documentclass[cancers,article,submit,pdftex,moreauthors]{Definitions/mdpi}

\usepackage{longtable} 

\makeatletter
\renewcommand{\linenumbers}{}   
\renewcommand{\leftcolDates}{}  
\renewcommand{\cright}{%
	\fontdimen2\font=1.2pt
	\raggedright\textbf{Copyright: }\copyright{} {\@ \the\year} by the \@authornum.
	This preprint is distributed under the terms and conditions of the
	\href{https://creativecommons.org/licenses/by/4.0/}%
	{Creative Commons Attribution (CC~BY) license}.%
}
\makeatother

\fancypagestyle{plain}{%
	\fancyhf{}%
	\fancyhead{\null\vspace{8pt}}%
}

\firstpage{1}
\pubvolume{1}
\issuenum{1}
\articlenumber{0}
\pubyear{2026}
\copyrightyear{2026}
\datereceived{ }
\daterevised{ }
\dateaccepted{ }
\datepublished{ }

\Title{Searching for Robust Augmentations to Improve Out-of-Domain Generalization in Dermoscopic Skin Cancer Classification}

\Author{Alexander~Kozachok $^{1,}$*\orcidA{}, Ilya~Latyshev $^{1}$\orcidB{}, Evgeny~Karpulevich $^{1}$\orcidC{}, Elena~Kozachok $^{1}$\orcidD{}, Egor~Ushakov $^{1}$\orcidF{} and Oleg~Samovarov $^{1}$\orcidE{}}

\AuthorNames{A. Kozachok, I. Latyshev, E. Karpulevich, E. Kozachok, E. Ushakov and O. Samovarov}

\address{%
$^{1}$ \quad Trusted AI Research Center, Russian Academy of Sciences, 109004 Moscow, Russia; latyshev@tairc.ru (I.L.); karpulevich@tairc.ru (Ev.K.); e.kozachok@tairc.ru (El.K.); catsmeowmeoww@tairc.ru (E.U.); samov@tairc.ru (O.S.)}

\corres{Correspondence: a.kozachok@tairc.ru}

\simplesumm{Automated analysis of dermoscopic images is increasingly used to decide which skin lesions need urgent specialist review. A model that performs well on images from one clinic often fails on images from another, because dermatoscopes, illumination, colour processing and imaging artifacts differ. We asked which data augmentations---random image transformations applied during training---make a classifier less dependent on such source-specific cues. Single transformations, colour combinations and composite policies were ranked on a small development set, and the winning policy was then measured on 8073 held-out images from archives that took no part in that ranking and that share no lesion and no contributing institution with the training data. The selected policy improved discrimination by 0.039 ROC-AUC on that set, consistently across four independent training runs, while in-domain accuracy was preserved. Put in clinical terms, at a fixed proportion of missed cancers the augmented model sent about a quarter fewer harmless lesions for unnecessary specialist review. Augmentation policies are a simple and reproducible step that should be tuned before the clinical validation of skin-cancer triage systems.}

\abstract{\textbf{Background/Objectives:} Dermoscopic skin-lesion classifiers lose accuracy when images arrive from a new clinic or a new device. We asked which data augmentations reduce that loss, and we measured the effect under a protocol that keeps policy selection separate from policy evaluation. \textbf{Methods:} A ConvNeXt-Large binary malignant-versus-non-malignant classifier was trained on six dermoscopic sources (25,903 images); HAM10000 and ISIC~2016--2020 were held out of training entirely. Single augmentations, photometric combinations and eleven composite policies were ranked on a development split of 1511 held-out images. The winning policy was then evaluated on a confirmation set of 8073 held-out images that took no part in that ranking and from which we removed every image sharing a lesion identifier with the training data and every image contributed by an institution represented in training. Both policies were retrained with four random seeds each and compared with an exact permutation test. \textbf{Results:} The \emph{mix} policy raised confirmation-set ROC-AUC from 0.787 to 0.826 ($+0.039$; per-seed ranges 0.772--0.797 and 0.815--0.840, non-overlapping; exact permutation $p=0.029$), with the same direction on each contributing source. At matched sensitivity the gain is larger in clinical terms: specificity rose from 0.612 to 0.713 at a sensitivity of 0.80, and from 0.284 to 0.397 at a sensitivity of 0.95. In-domain ROC-AUC was preserved (0.938 to 0.941). On an independent clinical cohort acquired with a different device at a different institution (472 images, 22 malignant), performance was maintained (0.934 versus 0.930). \textbf{Conclusions:} Augmentations that model the physical causes of domain shift improve cross-source transfer at no cost to in-domain accuracy, and the improvement survives a selection-disjoint, contamination-free evaluation.}


\begin{document}

\section{Introduction}

The incidence of malignant skin neoplasms continues to rise, and early detection remains a key determinant of treatment strategy and prognosis. If current trends persist, the burden of cutaneous melanoma may increase substantially by 2040~\citep{ref-arnold}. Dermoscopy improves diagnostic accuracy over naked-eye examination, but the size of that improvement depends on the experience of the clinician and on the examination protocol~\citep{ref-kittler}. It is at primary screening that the decision is made whether a lesion requires biopsy, further examination, or surveillance, so this is where automated support is most useful.

Deploying deep learning here is complicated by domain shift. Images of the same pathology acquired in different clinics or with different dermatoscopes differ in colour distribution, contrast, artifact contamination and background structure. Convolutional networks reached dermatologist-level classification on curated data~\citep{ref-esteva}, but high internal accuracy does not guarantee robustness on an independent source; the ISIC~2019 Grand Challenge showed that statistical shift, novel diagnostic categories and image artifacts all degrade real-world reliability~\citep{ref-combalia}. Non-biological correlations aggravate the problem. Markers, stickers, hair, illumination and white balance can co-occur with the target class and be learned in its place. This is shortcut learning~\citep{ref-geirhos}, and in dermoscopy it is a documented and specific threat~\citep{ref-nauta}.

This work searches for augmentations that make a classifier less sensitive to such differences, and measures the result on sources excluded from training. Rather than compare architectures, we treat the data pipeline as the object of study: augmentation expands the training distribution and acts as a practical regularizer~\citep{ref-shorten}. The intended use is not autonomous diagnosis but case prioritization---flagging lesions that warrant attention, additional diagnostics or biopsy---and remaining reliable when images arrive from a new institution.

\subsection{Related Work and Research Gap}

Most dermoscopic classification studies evaluate within a single dataset, using an internal split or cross-validation. That protocol compares architectures well but overestimates transferability. The ISIC~2016--2018 challenges standardized such evaluation~\citep{ref-gutman,ref-codella2018,ref-codella2019}, and analysis of the ISIC~2016--2020 releases showed that duplicates and source heterogeneity make lesion-level splitting and explicit overlap control mandatory~\citep{ref-cassidy}. Sources of shift in dermatology include population, prevalence, lesion site, capture device and preprocessing; HAM10000 and Derm7pt are widely used precisely because they expose these differences~\citep{ref-tschandl,ref-kawahara}.

Augmentation itself has been studied here before. Perez et al.\ compared thirteen configurations of geometric and photometric transformations and found that augmentation choice can matter more than backbone choice~\citep{ref-perez}; Valle et al.\ reached a similar conclusion about design factors in skin-lesion pipelines~\citep{ref-valle}. Both evaluate within ISIC challenge splits, so their rankings are in-domain rankings, and neither isolates a source-level held-out test---the regime in which photometric augmentation should pay off. A parallel line learns the policy instead: AutoAugment and RandAugment optimize against validation accuracy~\citep{ref-cubuk}, and \emph{mixup} regularizes by interpolating images and labels~\citep{ref-zhang}. These are strong general-purpose baselines, but they optimize for the validation distribution, which under source-level shift is the wrong target, and they do not identify \emph{which class} of transformation produces any robustness gained.

A third line explains why robustness here is fragile. Bissoto et al.\ showed that skin-lesion classifiers retain much of their accuracy when the lesion is occluded, implying heavy reliance on background and artifact cues~\citep{ref-bissoto}; Nauta et al.\ catalogued specific shortcuts such as coloured patches and rulers~\citep{ref-nauta}. Independently curated sets confirm the consequences: performance drops on the diverse clinical images of DDI~\citep{ref-daneshjou} and varies with skin tone on Fitzpatrick17k~\citep{ref-groh}. This literature diagnoses the problem rather than prescribing a training-time recipe. Domain-adversarial training~\citep{ref-ganin} and test-time adaptation~\citep{ref-wang-tent} are more powerful in principle but require source labels during training or unlabeled target data at deployment; an augmentation policy requires neither.

Table~\ref{tab:related} places this study against that background. The gap is specific: augmentation studies in dermoscopy rank policies in-domain, domain-shift studies document failure without prescribing a policy, and automated-augmentation methods optimize an objective misaligned with source-level transfer. We address that intersection with a systematic search evaluated on a source-level held-out test, under a protocol that separates the images used to choose the policy from the images used to measure it.

\begin{table}[H]
\caption{Positioning of this work relative to the closest prior studies. ``Source-level OOD'' means that entire acquisition sources are held out of training, as opposed to a random or challenge-provided split. ``Stat.\ testing'' refers to confidence intervals or significance tests on the reported differences.\label{tab:related}}
\begin{adjustwidth}{-\extralength}{0cm}
\small
\begin{tabularx}{\fulllength}{Xlcccc}
\toprule
\textbf{Study} & \textbf{Focus} & \textbf{Aug.\ search} & \textbf{Source-level OOD} & \textbf{Stat.\ testing} & \textbf{Multi-seed}\\
\midrule
Perez et al.~\citep{ref-perez} & Augmentation for skin lesions & Yes & No & No & No\\
Valle et al.~\citep{ref-valle} & Design factors, skin lesions & Partial & No & Partial & Yes\\
Cubuk et al.~\citep{ref-cubuk} & Learned augmentation policy & Automated & No & No & No\\
Bissoto et al.~\citep{ref-bissoto} & Bias/shortcut diagnosis & No & No & No & No\\
Nauta et al.~\citep{ref-nauta} & Shortcut correction & No & No & Partial & No\\
Daneshjou et al.~\citep{ref-daneshjou} & External clinical validation & No & Yes & Yes & No\\
Groh et al.~\citep{ref-groh} & Skin-tone generalization & No & Yes & Partial & No\\
\midrule
\textbf{This work} & \textbf{Augmentation for domain shift} & \textbf{Yes} & \textbf{Yes} & \textbf{Yes} & \textbf{Yes}\\
\bottomrule
\end{tabularx}
\end{adjustwidth}
\end{table}

\section{Materials and Methods}

\subsection{Study Design}

The task is binary classification into malignant and non-malignant, defined by the label mapping of Section~\ref{sec:data}. Two things must be kept apart. The \emph{computational} task is that dichotomy. The \emph{intended use} is narrower: the output prioritizes cases for clinical assessment and does not certify a lesion as safe. The distinction matters because the non-malignant class pools genuinely benign lesions with premalignant ones, actinic keratosis in particular, so a negative prediction does not imply that observation alone is adequate management. Three levels of decision were compared: no extended augmentation search (baseline), individual transformations, and composite policies assembled from photometric, geometric and artifact-modelling operations.

The primary metric is ROC-AUC. It is threshold-independent and relatively insensitive to prevalence, which matters because the evaluation sets differ substantially in class balance.

\subsection{Materials and Datasets}\label{sec:data}

Two kinds of material were used. The first is a set of public research collections that supply all training data and all out-of-domain evaluation: the ISIC Archive datasets BCN20000, Derm12345, HAM10000, ISIC~2016--2020, HIBA, BALD and MILK10k, together with Derm7pt, which is not part of the ISIC Archive. The second is \emph{Melanoscope}, a closed clinical collection assembled and owned by the authors, used only to evaluate already-trained models.

ISIC Archive images were filtered by the \texttt{diagnosis\_3} field, the most specific level of the archive's diagnostic hierarchy populated consistently across all collections used here. Three criteria were applied. Only lesions whose \texttt{diagnosis\_3} maps unambiguously to one side of the dichotomy were kept; entries with missing, purely hierarchical or indeterminate labels were discarded. Only \texttt{image\_type} = \texttt{dermoscopic} images were retained, so the study measures shift between dermatoscopes rather than between imaging modalities. Categories with too few qualifying images to support a stable per-source estimate were dropped. The resulting sixteen categories were collapsed into nine label groups. Only images with a unique \texttt{isic\_id} were kept. From Derm7pt, melanosis cases were excluded. This filtering explains why the counts reported here are smaller than the nominal sizes of the source collections. The sources were chosen to maximize inter-domain variability across clinics, devices and protocols rather than to maximize sample size.

Melanoma, including invasive, \emph{in situ} and metastatic forms, basal cell carcinoma and squamous cell carcinoma were assigned to the malignant class; nevi, seborrheic and pigmented benign keratoses, lichen-planus-like keratoses, actinic keratoses, dermatofibromas, solar lentigines and haemangiomas form the non-malignant class. We use ``non-malignant'' rather than ``benign'' deliberately: the negative class pools genuinely benign lesions with precursor entities, most notably actinic keratosis. The clinical question modelled is therefore ``does this lesion require specialist referral?'' rather than ``is this lesion normal?''. Mappings are given in Table~\ref{tab:map-isic} and Table~\ref{tab:map-derm7pt}.

\begin{table}[H]
\caption{Mapping of ISIC Archive diagnoses to the binary malignant/non-malignant classes.\label{tab:map-isic}}
\begin{tabularx}{\textwidth}{Xc}
\toprule
\textbf{Diagnosis} & \textbf{Class}\\
\midrule
Melanoma & Malignant\\
Basal cell carcinoma & Malignant\\
Squamous cell carcinoma & Malignant\\
Nevus & Non-malignant\\
Keratosis & Non-malignant\\
Solar or actinic keratosis & Non-malignant\\
Dermatofibroma & Non-malignant\\
Lentigo & Non-malignant\\
Hemangioma & Non-malignant\\
\bottomrule
\end{tabularx}
\end{table}

\begin{table}[H]
\caption{Mapping of Derm7pt diagnoses to the binary malignant/non-malignant classes.\label{tab:map-derm7pt}}
\begin{tabularx}{\textwidth}{Xc}
\toprule
\textbf{Diagnosis} & \textbf{Class}\\
\midrule
Basal cell carcinoma & Malignant\\
Melanoma (less than 0.76 mm) & Malignant\\
Melanoma (0.76 to 1.5 mm) & Malignant\\
Melanoma (more than 1.5 mm) & Malignant\\
Melanoma metastasis & Malignant\\
Melanoma (in situ) & Malignant\\
Melanoma & Malignant\\
Blue nevus & Non-malignant\\
Clark nevus & Non-malignant\\
Combined nevus & Non-malignant\\
Congenital nevus & Non-malignant\\
Dermal nevus & Non-malignant\\
Dermatofibroma & Non-malignant\\
Lentigo & Non-malignant\\
Recurrent nevus & Non-malignant\\
Reed or Spitz nevus & Non-malignant\\
Seborrheic keratosis & Non-malignant\\
Vascular lesion & Non-malignant\\
\bottomrule
\end{tabularx}
\end{table}

Six sources---BCN20000, Derm12345, Derm7pt, HIBA, BALD and MILK10k---form the in-domain partition. HAM10000 and ISIC~2016--2020 were held out of training entirely and supply all out-of-domain evaluation. Every source was partitioned into train/validation/test in an approximately 70/21/9 ratio, with two documented exceptions. For Derm7pt we used the official partition distributed with the dataset (384/185/383 images after restricting to dermoscopic images and excluding melanosis), which keeps our in-domain results comparable with published work on that benchmark. For ISIC~2019 the partition is 2524/265/110; since none of its images are used for fitting, this ratio affects only how its images are distributed between the two evaluation levels defined in Section~\ref{sec:protocol}.

Splits were made at the lesion level wherever the archive supplies a lesion identifier, which covers BCN20000, HIBA, MILK10k, HAM10000, ISIC~2019 and ISIC~2020; within each of these sources no test image depicts a lesion that also appears in its own training or validation split. BALD, Derm12345 and Derm7pt are distributed without a \texttt{lesion\_id} field, so for those three sources the split is at the image level and repeated imaging of the same lesion cannot be excluded within the in-domain partition. This affects the in-domain figures only; every out-of-domain source carries lesion identifiers.

The external \emph{Melanoscope} collection was acquired with devices, populations and protocols not represented in training. It is the only evaluation set in this study that originates outside the ISIC Archive and its contributing institutions. It was de-identified before analysis and analysed retrospectively (see the Institutional Review Board Statement). It is used exclusively for the external evaluation of Section~\ref{sec:external}, never for training, for the augmentation search, or for tuning the decision threshold.

\subsection{Evaluation Protocol}\label{sec:protocol}

An augmentation policy is a top-level hyperparameter. If it is chosen on the same images used to report its effect, the reported effect is optimistically biased, both by selection and by the winner's curse over the candidate set. The evaluation is therefore organized into three levels, and each number in this paper is labelled with the level it comes from.

The \emph{development set} is the pooled test split of the held-out sources: HAM10000, ISIC~2019 and ISIC~2020. All eleven candidate policies were ranked on it (Section~\ref{sec:selection}), and the decision thresholds of Section~\ref{sec:impl} were tuned before it. It is a model-selection set, and the differences measured on it are reported as such.

The \emph{confirmation set} is the remainder of the held-out collections: the ISIC~2019 and ISIC~2020 images outside their test splits, together with ISIC~2016. No image in it took any part in ranking the policies. Splits were made uniformly across all sources before the search began, and the search touched only the test portion of the held-out sources; this is what makes a disjoint confirmation set available after the fact. It carries the primary result of the study. HAM10000 contributes no images to this level, because only its test split entered the evaluation pool at all.

The \emph{external cohort} is Melanoscope, which shares neither an archive nor a contributing institution with any other data used here.

The two public levels are also disjoint at the lesion level, not merely at the image level: the intersection of their \texttt{lesion\_id} sets is empty, so no lesion contributes an image to both the set that chose the policy and the set that measures it. The same holds within each held-out source, where no test-split image depicts a lesion appearing in that source's own training or validation portion.

Two contamination filters are applied to both public levels before any metric is computed. First, MILK10k and HAM10000 partly follow the same physical lesions over time and share 967 archive-level \texttt{lesion\_id} values; 158 HAM10000 test images depict a lesion that also appears, as a different image from a different visit, in the MILK10k training or validation split. All 158 are removed. Second, 179 ISIC~2020 images are attributed to the Department of Dermatology, Hospital Clínic de Barcelona, the institution that contributed BCN20000 to the in-domain partition; although they share no \texttt{lesion\_id} with BCN20000, they are not institutionally independent of the training data, and all 179 are removed. After both filters the evaluation pool contains 9584 of the original 9921 images. We verified separately that no image-level overlap exists anywhere: the intersection of \texttt{isic\_id} sets between every in-domain and every held-out collection is empty, no \texttt{isic\_id} occurs twice within a collection, and the held-out collections are mutually disjoint at both the image and the lesion level. Results on the unfiltered pool are reported in Appendix~\ref{app:dist} for comparison; removing the contaminated images widens the measured gap rather than narrowing it.

Table~\ref{tab:dist} summarizes the resulting partitions; a per-source breakdown is given in Appendix~\ref{app:dist}.

\begin{table}[H]
\caption{Distribution of images across partitions. The in-domain partitions are pooled over the six in-domain sources. The development and confirmation sets are the two levels of the held-out evaluation defined in Section~\ref{sec:protocol}, after the two contamination filters. The external cohort is the Melanoscope clinical collection.\label{tab:dist}}
\begin{adjustwidth}{-\extralength}{0cm}
\small
\begin{tabularx}{\fulllength}{Xccccccc}
\toprule
& \multicolumn{3}{c}{\textbf{In-domain}} & \multicolumn{2}{c}{\textbf{Held-out (public)}} & \textbf{External}\\
\cmidrule(lr){2-4}\cmidrule(lr){5-6}\cmidrule(lr){7-7}
\textbf{Class} & \textbf{Train} & \textbf{Validation} & \textbf{Test} & \textbf{Development} & \textbf{Confirmation} & \textbf{Melanoscope}\\
\midrule
Non-malignant & 15{,}972 & 4737 & 2156 & 1319 & 7172 & 450\\
Malignant & 9931 & 3058 & 1441 & 192 & 901 & 22\\
\midrule
\textbf{Total} & \textbf{25{,}903} & \textbf{7795} & \textbf{3597} & \textbf{1511} & \textbf{8073} & \textbf{472}\\
\textbf{Prevalence} & 0.38 & 0.39 & 0.40 & 0.13 & 0.11 & 0.05\\
\bottomrule
\end{tabularx}
\end{adjustwidth}
\end{table}

\subsection{Evidence of Domain Shift}\label{sec:domainshift}

\begin{figure}[H]
\centering
\includegraphics[width=\textwidth]{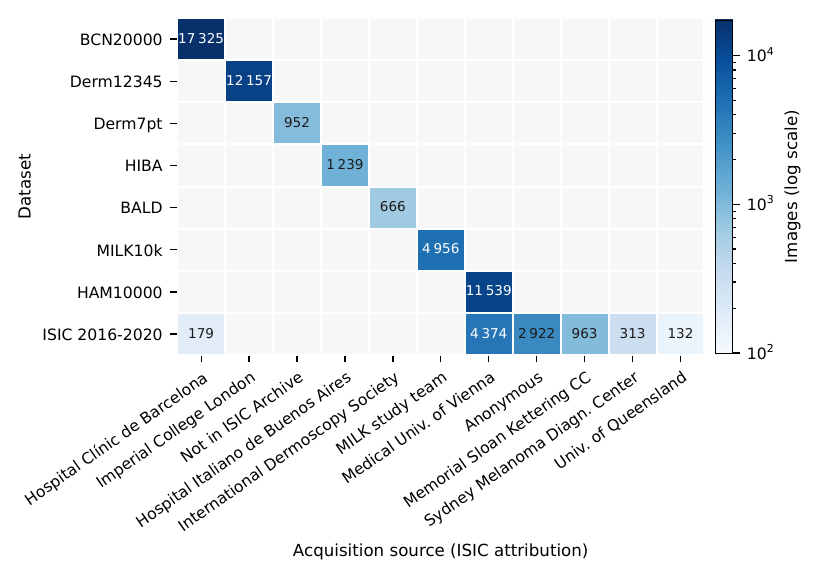}
\caption{Distribution of ISIC \texttt{attribution} values across the datasets. Counts are taken from the split files actually used in this study, so they agree with Table~\ref{tab:dist-source}; the colour scale is logarithmic and empty combinations are left blank. Every dataset except the ISIC~2016--2020 group maps onto a single acquisition source. The 179 ISIC~2020 images attributed to Hospital Clínic de Barcelona are the institutional overlap with BCN20000 removed by the second contamination filter of Section~\ref{sec:protocol}. Derm7pt is not part of the ISIC Archive and has no ISIC attribution.\label{fig:sources}}
\end{figure}

The ISIC \texttt{attribution} field records the collecting clinic or team. Its distribution across the datasets is concentrated: several sub-collections, in particular ISIC~2016--2020 and HAM10000, come predominantly from a small number of sources including the Medical University of Vienna (Figure~\ref{fig:sources}). This matters for reliability, because a model can use the source as a proxy for the diagnosis whenever classes and sources are unevenly distributed.

\begin{figure}[H]
\centering
\includegraphics[width=\textwidth]{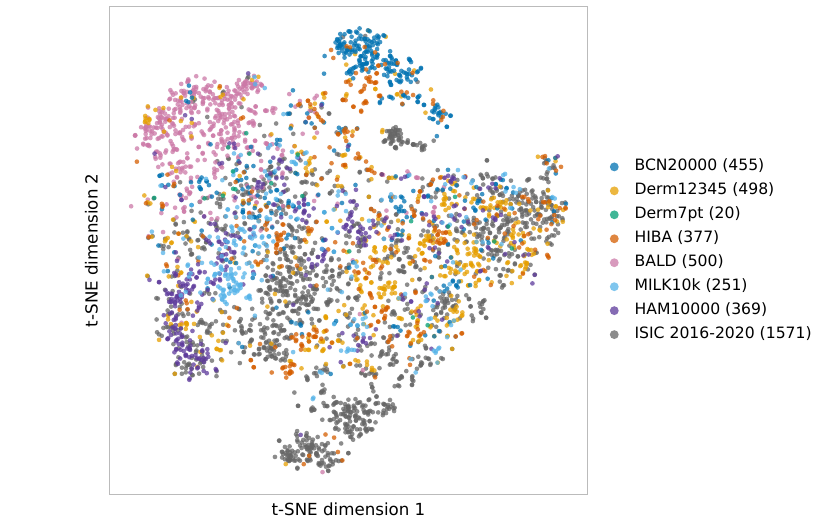}
\caption{t-SNE visualization of test-image embeddings from ImageNet-pretrained ConvNeXt-Large, coloured by dataset ($N$ per dataset in the legend). Partial grouping by acquisition source indicates source-specific features. t-SNE axes carry no units and are shown without ticks.\label{fig:imagenet-tsne}}
\end{figure}

Features from the penultimate layer of an ImageNet-pretrained ConvNeXt-Large cluster partially by source in a t-SNE projection, which is consistent with source-specific visual patterns; t-SNE is used here as a diagnostic for embedding structure, not as a statistical test~\citep{ref-vandermaaten} (Figure~\ref{fig:imagenet-tsne}).

\begin{figure}[H]
\centering
\includegraphics[width=\textwidth]{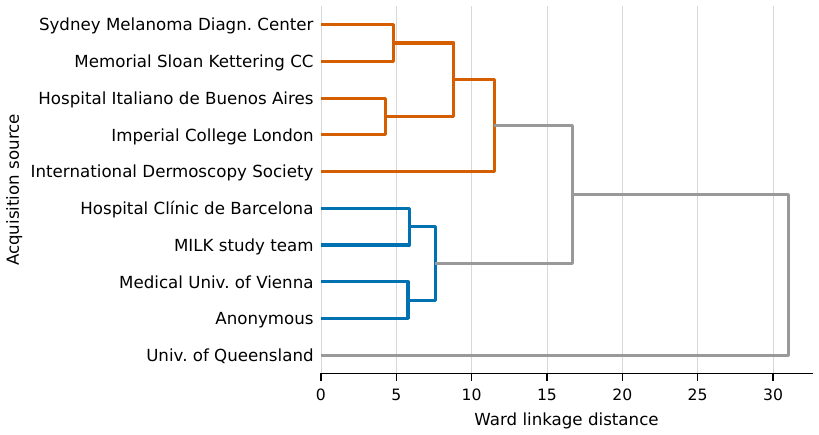}
\caption{Dendrogram of the clustering of data-collection sources by colour distributions and inter-histogram distance metrics.\label{fig:dendrogram}}
\end{figure}

K-means and hierarchical agglomerative clustering of the sources, using distance matrices built from the symmetrized Kullback--Leibler divergence, the Jensen--Shannon distance and the Wasserstein distance over RGB channels, separate the Medical University of Vienna and one anonymous source into their own cluster (Figure~\ref{fig:dendrogram}). Differences in dermatoscope, illumination, colour correction and white balance are the most plausible explanation, and this is why the search below emphasizes photometric augmentation.

\subsection{Model and Implementation}\label{sec:impl}

The backbone is ConvNeXt-Large, initialized from ImageNet-pretrained weights (\texttt{ConvNeXt\_Large\_Weights.DEFAULT}) with a single-logit binary head and a sigmoid output. ConvNeXt keeps a convolutional inductive bias while adopting design principles from Vision Transformers, which makes it both a strong classifier and a natural fit for classical image augmentation~\citep{ref-liu}. A single backbone limits architecture-level conclusions; cross-checking on a Vision Transformer and an EfficientNet is left for future work.

Images were resized to $224\times224$ and normalized with dataset-specific channel means $(0.658, 0.542, 0.499)$ and standard deviations $(0.237, 0.217, 0.219)$ estimated on the training data. The network was optimized with AdamW (learning rate $8\times10^{-5}$, $\beta=(0.9, 0.999)$, weight decay $0.01$) under a focal loss, with a StepLR schedule (factor $0.8$ every 1500 steps), batch size 16, for 15 epochs. The best checkpoint was selected by macro ROC-AUC on the validation set. The decision threshold used for operating-point metrics was tuned on the in-domain validation split by maximizing the Youden index, giving $0.476$ for the baseline and $0.455$ for \emph{mix}; it was tuned separately for each policy and each run, always on validation data, and no held-out or external image took part in choosing it. The baseline is the same backbone, preprocessing and optimization trained without the augmentation search---resizing, normalization and horizontal/vertical/transpose flips only---and the policies are applied on top of that pipeline. Augmentations were implemented with \texttt{albumentations} 2.0.8. Experiments were tracked with MLflow.

\subsection{Augmentation Search Space}

Photometric transformations change colour and intensity and target variability from device, illumination, white balance and skin coloration: ColorJitter, PlanckianJitter, HueSaturationValue, ToGray, ChromaticAberration, RandomGamma, CLAHE, RandomBrightnessContrast and HEStain. The intended effect is that the model should depend less on absolute RGB values and local colour shifts, which reflect the source more than the biology of the lesion.

HEStain originates as a stain perturbation for haematoxylin--eosin histopathology, where stain colour augmentation is well established for robustness across scanners and laboratories~\citep{ref-macenko,ref-tellez}. It is not a native dermoscopic transformation, and we do not use it as a physical model of dermoscopy. Using the Macenko decomposition~\citep{ref-macenko}, it factors an image into stain-like density channels and randomly reweights them, producing strong but structure-preserving chromatic variation. It therefore acts as a domain-randomization operator on exactly the channel statistics---overall colour cast, contrast between pigmented and background regions---that Section~\ref{sec:domainshift} found to differ most across dermatoscopes.

Geometric and artifact-modelling transformations regularize spatial features and increase invariance to position, scale, small deformation and partial occlusion: Transpose, VerticalFlip, HorizontalFlip, Rotate, MotionBlur, MedianBlur, GaussianBlur, GaussNoise, OpticalDistortion, GridDistortion, CoarseDropout, ShiftScaleRotate, ElasticTransform, Affine and GridDropout. Aggressive geometric and occlusion transformations can destroy clinically meaningful structure, so their effect must be read against both evaluation levels rather than a single aggregate score.

The hypothesis is that photometric augmentation dominates for cross-source transfer, because that is where spectral and colour characteristics differ, while geometric augmentation mainly improves robustness to local variation and limits overfitting to incidental spatial correlations. This follows the general logic of robust augmentation: expand the training distribution to cover variation expected at deployment~\citep{ref-hendrycks}.

\subsection{Experimental Setup}\label{sec:setup}

Single augmentations, combinations of photometric transformations, and composite sets inspired by competition practice were evaluated in turn. Eleven composite policies were compared: \emph{simple}, \emph{prime}, \emph{strong\_hestain}, \emph{kaggle}, \emph{kaggle\_simple}, \emph{kaggle\_strong}, \emph{kaggle\_strong\_hestain}, \emph{mix}, \emph{combine\_all}, \emph{low\_intensity\_prime} and \emph{low\_intensity\_kaggle\_strong}. Their exact compositions are in Appendix~\ref{app:aug}.

\subsection{Statistical Analysis}\label{sec:stats}

Two sources of variability are distinguished throughout, because they answer different questions.

Training variance is the primary one. Both the baseline and \emph{mix} were retrained with four random seeds each---baseline seeds 42, 1, 10, 20 and \emph{mix} seeds 2, 5, 15, 30---with the data split and all hyperparameters held fixed. Because the two policies use different seed sets the runs cannot be paired, so the groups are compared as independent samples with an exact two-sided permutation test on the difference of group means, enumerating all $\binom{8}{4}=70$ assignments. With four runs per group the smallest attainable $p$-value is $2/70=0.029$, reached exactly when the two groups do not overlap.

Test-set sampling variability is secondary and is reported for one reference checkpoint pair, for comparability with the many single-run studies in this literature. For that pair we report ROC-AUC with a 95\% stratified bootstrap confidence interval (2000 resamples), the paired difference with its own bootstrap interval, and the two-sided DeLong test for correlated ROC curves. A DeLong $p$-value speaks only to whether two already-trained models differ on a given sample; it says nothing about whether the difference survives retraining, which is why the seed-level test carries the conclusion.

Operating-point metrics use the validation-tuned thresholds of Section~\ref{sec:impl} and are reported as sensitivity, specificity, balanced accuracy and area under the precision--recall curve. No correction is applied for the eleven-way policy search; the confirmation set exists precisely so that the reported effect does not depend on that correction. All statistics were recomputed from retained per-image predictions with a single script, which is released with the manuscript.

\section{Results}

\subsection{Single Augmentations and Colour Combinations}

Screening of single augmentations on the development set identified ColorJitter, ChromaticAberration, CLAHE, GridDistortion, GridDropout, HueSaturationValue, OpticalDistortion and PlanckianJitter as the most useful transformations for out-of-domain transfer. Operations that change colour, contrast and local geometry dominate this list---precisely the factors that differ between dermatoscopes and capture protocols. On the in-domain test the useful set is different: Affine, ColorJitter, GridDropout, HEStain, PlanckianJitter and Rotate. Affine and Rotate also delayed the onset of overfitting, indicating a regularizing effect. The usefulness of an augmentation therefore depends on the evaluation domain: a transformation may barely move in-domain accuracy yet pay off on transfer.

Eleven combinations of photometric operators were then evaluated, each a fixed pairing or triple of ColorJitter, PlanckianJitter, HueSaturationValue and HEStain (Appendix~\ref{app:aug}, Table~\ref{tab:color}). Sets 1, 2, 3, 4, 7, 8 and 11 improved out-of-domain performance, and sets 1--4 improved the in-domain test as well. Effectiveness thus depends sharply on composition even within this narrow group: adding a colour operation is not automatically beneficial, and intensity, application probability and compatibility with the other operators all matter.

\subsection{Policy Selection on the Development Set}\label{sec:selection}

Table~\ref{tab:results} reports the change in ROC-AUC relative to the baseline for the eleven composite policies, measured on the development set. These are the numbers that selected the policy. Each policy and the baseline were trained once in this campaign, so the entries are single-run point estimates; they rank candidates and are not intended as effect sizes.

\begin{table}[H]
\caption{Change in ROC-AUC relative to the baseline for the composite augmentation policies, measured on the development set. Positive values indicate improvement. The best value in each column is shown in bold. All entries come from the policy-search campaign, in which each policy and the baseline were trained once; the deltas are single-run point estimates used to rank candidates, and are not directly comparable with the separately trained checkpoints analysed in Section~\ref{sec:confirm} and Section~\ref{sec:seeds}.\label{tab:results}}
\begin{tabularx}{\textwidth}{Xcc}
\toprule
\textbf{Augmentation Set} & \textbf{$\Delta$ ROC-AUC In-Domain} & \textbf{$\Delta$ ROC-AUC Development}\\
\midrule
low\_intensity\_prime & 0.0004 & 0.0099\\
low\_intensity\_kaggle\_strong & $-$0.0005 & $-$0.0030\\
combine\_all & 0.0040 & 0.0060\\
kaggle & 0.0051 & 0.0173\\
kaggle\_simple & \textbf{0.0072} & 0.0129\\
kaggle\_strong\_hestain & 0.0026 & 0.0303\\
mix & 0.0061 & \textbf{0.0332}\\
prime & 0.0001 & 0.0244\\
simple & $-$0.0019 & 0.0234\\
strong\_hestain & $-$0.0019 & 0.0248\\
\bottomrule
\end{tabularx}
\end{table}

The largest out-of-domain gain belongs to \emph{mix} ($+0.0332$), followed by \emph{kaggle\_strong\_hestain} ($+0.0303$) and \emph{strong\_hestain} ($+0.0248$). In-domain the best is \emph{kaggle\_simple} ($+0.0072$), then \emph{mix} ($+0.0061$) and \emph{kaggle} ($+0.0051$). \emph{Mix} is therefore the most balanced configuration: it achieves the maximum external gain while keeping a positive in-domain effect, and it was selected on that basis.

The trade-off visible in the table is the characteristic one of domain generalization. Several policies give up a little in-domain accuracy for a substantially larger external gain---\emph{simple} and \emph{strong\_hestain} lose $0.0019$ in-domain and gain $+0.0234$ and $+0.0248$ respectively---which is what one expects when a model discards locally useful features of the training source in exchange for transfer.

One point of arithmetic should be settled before the next section, because the deltas there are larger than the $+0.0332$ reported here. The difference is not in the data or the metric but in the checkpoints. Table~\ref{tab:results} comes from the search campaign, in which every candidate and its baseline were trained once. Sections~\ref{sec:seeds} and~\ref{sec:confirm} use separately trained baseline and \emph{mix} checkpoints, for which the per-image predictions were retained. Evaluated on the identical 1685 images of the unfiltered screening split, the search-campaign pair gives $+0.033$ and the retained pair gives $+0.058$, while the eight runs of Section~\ref{sec:seeds} produce pairwise gaps spanning $+0.028$ to $+0.065$ on the same images. Both values lie inside the ordinary spread of two independently trained models. A single-run margin should therefore not be read to three decimal places, which is exactly why the policy ranking above is treated as a ranking and the effect size is taken from the multi-seed analysis.

\subsection{Confirmation on Held-Out Images}\label{sec:seeds}

The primary result is the behaviour of \emph{mix} on the confirmation set, whose 8073 images played no part in the ranking above and contain no lesion and no institution shared with the training data. Table~\ref{tab:seeds} reports ROC-AUC across four training runs per policy.

\begin{table}[H]
\caption{Primary analysis: ROC-AUC across four random seeds per policy (baseline seeds 42/1/10/20; \emph{mix} seeds 2/5/15/30), with the exact two-sided permutation $p$-value over all 70 group assignments. With four runs per group the smallest attainable $p$-value is $0.029$, reached when the per-seed ranges do not overlap. Rows reaching it are shown in bold.\label{tab:seeds}}
\begin{adjustwidth}{-\extralength}{0cm}
\small
\begin{tabularx}{\fulllength}{Xcccccc}
\toprule
\textbf{Evaluation Set} & \textbf{\textit{N}} & \textbf{Baseline (mean~$\pm$~SD)} & \textbf{Range} & \textbf{Mix (mean~$\pm$~SD)} & \textbf{Range} & \textbf{\textit{p}}\\
\midrule
\textbf{Confirmation} & 8073 & $0.787 \pm 0.011$ & 0.772--0.797 & $\mathbf{0.826 \pm 0.012}$ & 0.815--0.840 & \textbf{0.029}\\
Development & 1511 & $0.839 \pm 0.010$ & 0.827--0.850 & $0.887 \pm 0.005$ & 0.884--0.895 & \textbf{0.029}\\
In-domain test & 3597 & $0.938 \pm 0.003$ & 0.935--0.941 & $0.941 \pm 0.001$ & 0.940--0.942 & 0.086\\
External (Melanoscope) & 472 & $0.934 \pm 0.015$ & 0.917--0.950 & $0.930 \pm 0.022$ & 0.907--0.953 & 0.71\\
\bottomrule
\end{tabularx}
\end{adjustwidth}
\end{table}

On the confirmation set \emph{mix} outperforms the baseline in every one of the eight runs. The per-seed ranges are 0.772--0.797 for the baseline and 0.815--0.840 for \emph{mix}, with no overlap, and the mean gap of $+0.039$ is more than three times the within-group standard deviation of either policy. The permutation test reaches its floor at $p = 0.029$. The effect is therefore not an artifact of one favourable initialization, and, because no image in this set was used to rank the policies, it is not an artifact of the selection either.

The improvement is not an artifact of pooling one large source with several small ones, although it is not equally strong in each. Broken down by source within the confirmation set, ROC-AUC rises from $0.784 \pm 0.013$ to $0.835 \pm 0.014$ on ISIC~2020 (5261 images; per-seed ranges 0.768--0.796 and 0.822--0.855, non-overlapping, $p = 0.029$) and from $0.752 \pm 0.021$ to $0.772 \pm 0.002$ on ISIC~2019 (2789 images; $+0.020$, $p = 0.057$). The ISIC~2019 ranges very nearly separate---the best baseline run reaches 0.771 against a worst \emph{mix} run of 0.770---so the direction is consistent there but the effect is smaller and does not reach the attainable floor of the test. ISIC~2016 contributes only 23 images and is not analysed separately. HAM10000 contributes no images to this level; on the development level, where it does appear, ROC-AUC rises from $0.886 \pm 0.018$ to $0.920 \pm 0.004$ with non-overlapping ranges.

In-domain accuracy is preserved rather than sacrificed: $0.938 \pm 0.003$ against $0.941 \pm 0.001$. The direction is favourable in all four runs, but with a gap of $+0.003$ the permutation test does not reach significance ($p = 0.086$), and we claim no more than the absence of a cost.

Ranking quality and precision quality do not move together, and the distinction matters at this prevalence. Area under the precision--recall curve on the confirmation set rises from $0.383 \pm 0.014$ to $0.395 \pm 0.008$ across the same eight runs, a gain of $+0.012$ that the permutation test does not resolve ($p = 0.20$). The reference checkpoint pair shows a larger AUPRC gap (Table~\ref{tab:confirm-op}), but the multi-seed spread does not support treating that gap as reproducible. The robust claim of this study is therefore about discrimination---the ordering of lesions by risk---and not about precision at a fixed operating point.

\subsection{Reference-Checkpoint Statistics}\label{sec:confirm}

Table~\ref{tab:confirm-auc} reports the same comparison for one retained checkpoint pair, with bootstrap confidence intervals and DeLong tests, so that the effect can be read against the single-run conventions of this literature.

\begin{table}[H]
\caption{Secondary analysis: baseline versus \emph{mix} for one retained checkpoint pair. ROC-AUC and $\Delta$AUC carry 95\% stratified bootstrap confidence intervals (2000 resamples); $p$ is the two-sided DeLong test for the paired difference. ``Prev.''\ is malignant prevalence. These intervals describe test-set sampling only and do not account for training variance, which is covered by Table~\ref{tab:seeds}.\label{tab:confirm-auc}}
\begin{adjustwidth}{-\extralength}{0cm}
\small
\begin{tabularx}{\fulllength}{Xcccccc}
\toprule
\textbf{Evaluation Set} & \textbf{\textit{N}} & \textbf{Prev.} & \textbf{Baseline AUC (95\% CI)} & \textbf{Mix AUC (95\% CI)} & \textbf{$\Delta$AUC (95\% CI)} & \textbf{\textit{p}}\\
\midrule
Confirmation & 8073 & 0.11 & 0.786 (0.771--0.800) & 0.832 (0.819--0.845) & \textbf{+0.047 (+0.037--+0.056)} & \textbf{<0.001}\\
Development & 1511 & 0.13 & 0.838 (0.802--0.873) & 0.895 (0.868--0.920) & \textbf{+0.057 (+0.039--+0.075)} & \textbf{<0.001}\\
In-domain test & 3597 & 0.40 & 0.935 (0.928--0.943) & 0.943 (0.936--0.950) & \textbf{+0.007 (+0.003--+0.012)} & \textbf{<0.001}\\
External (Melanoscope) & 472 & 0.05 & 0.916 (0.843--0.974) & 0.941 (0.878--0.993) & +0.026 ($-$0.007--+0.071) & 0.22\\
\bottomrule
\end{tabularx}
\end{adjustwidth}
\end{table}

\begin{figure}[H]
\centering
\includegraphics[width=\textwidth]{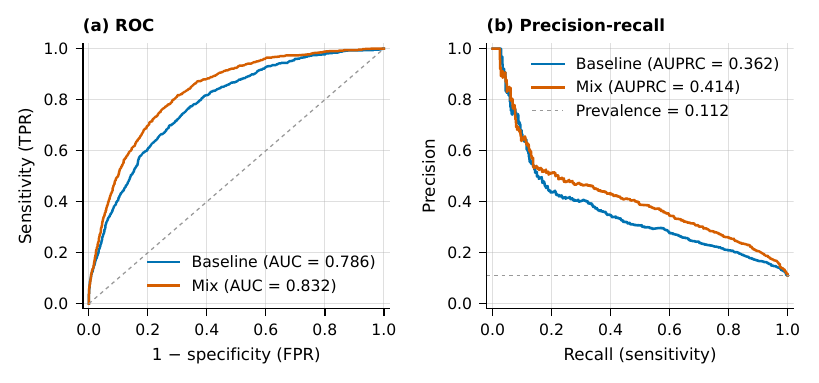}
\caption{Discrimination on the confirmation set ($N = 8073$, malignant prevalence $0.112$) for the baseline and the \emph{mix} policy: (\textbf{a})~receiver operating characteristic curves; (\textbf{b})~precision--recall curves, with the dashed line marking the malignant prevalence (the no-skill baseline). The \emph{mix} policy improves the area under the ROC curve from 0.786 to 0.832 and the area under the precision--recall curve from 0.362 to 0.414.\label{fig:roc-ood}}
\end{figure}

Figure~\ref{fig:roc-ood} shows the corresponding curves on the confirmation set. The \emph{mix} curve dominates the baseline across essentially the whole operating range, and the separation is clearest in the precision--recall view, which is the more informative one at a malignant prevalence of 0.11.

Repeated imaging of the same lesion makes an image-level bootstrap slightly optimistic. Resampling lesions rather than images widens the intervals only marginally and leaves the interval on the difference unchanged to three decimal places, so the correlation induced by repeat visits does not materially inflate the reported precision.

Operating-point metrics at the validation-tuned thresholds are given in Table~\ref{tab:confirm-op}. On the confirmation set \emph{mix} improves specificity from 0.920 to 0.941 and AUPRC from 0.362 to 0.414 at slightly higher sensitivity. In-domain the two models trade sensitivity for specificity at essentially unchanged balanced accuracy.

\begin{table}[H]
\caption{Operating-point metrics at the validation-tuned decision thresholds (baseline $0.476$, \emph{mix} $0.455$). Sens.~=~sensitivity, Spec.~=~specificity, Bal.~Acc.~=~balanced accuracy, AUPRC~=~area under the precision--recall curve.\label{tab:confirm-op}}
\begin{tabularx}{\textwidth}{Xlcccc}
\toprule
\textbf{Evaluation Set} & \textbf{Model} & \textbf{Sens.} & \textbf{Spec.} & \textbf{Bal. Acc.} & \textbf{AUPRC}\\
\midrule
\multirow{2}{*}{Confirmation} & Baseline & 0.365 & 0.920 & 0.643 & 0.362\\
 & Mix & 0.371 & 0.941 & 0.656 & 0.414\\
\midrule
\multirow{2}{*}{Development} & Baseline & 0.677 & 0.860 & 0.768 & 0.569\\
 & Mix & 0.615 & 0.939 & 0.777 & 0.658\\
\midrule
\multirow{2}{*}{In-domain test} & Baseline & 0.872 & 0.860 & 0.866 & 0.896\\
 & Mix & 0.861 & 0.874 & 0.868 & 0.911\\
\midrule
\multirow{2}{*}{External (Melanoscope)} & Baseline & 0.591 & 0.980 & 0.785 & 0.698\\
 & Mix & 0.818 & 0.967 & 0.892 & 0.796\\
\bottomrule
\end{tabularx}
\end{table}

\subsection{Operating Points Matched on Sensitivity}\label{sec:matched}

The thresholds of Table~\ref{tab:confirm-op} were fixed on in-domain validation data, where malignant prevalence is 0.39. Applied unchanged to a pool whose prevalence is 0.11, they land far down the sensitivity axis, which is why both models show sensitivity below 0.4 there. That is a property of the transferred threshold, not of the models, and it obscures the comparison: two classifiers evaluated at different points on their own ROC curves are not directly comparable in sensitivity and specificity.

We therefore also compare the policies at \emph{matched} sensitivity, setting each model's threshold on the confirmation set to reach the same recall and reading off the resulting specificity. This is the form in which a triage aid is normally specified---fix the miss rate that is clinically acceptable, then ask how many unnecessary referrals each model generates. Table~\ref{tab:matched} reports it across all four runs per policy.

\begin{table}[H]
\caption{Specificity at matched sensitivity on the confirmation set, across four runs per policy. For each run the threshold is set on that model alone to reach the target sensitivity, so the two policies are compared at the same miss rate. $p$ is the exact two-sided permutation test over the eight runs; rows reaching its attainable floor of 0.029 are shown in bold.\label{tab:matched}}
\begin{adjustwidth}{-\extralength}{0cm}
\small
\begin{tabularx}{\fulllength}{Xccccc}
\toprule
\textbf{Target sensitivity} & \textbf{Baseline spec.\ (mean~$\pm$~SD)} & \textbf{Range} & \textbf{Mix spec.\ (mean~$\pm$~SD)} & \textbf{Range} & \textbf{\textit{p}}\\
\midrule
0.80 & $0.612 \pm 0.029$ & 0.575--0.634 & $\mathbf{0.713 \pm 0.028}$ & 0.681--0.736 & \textbf{0.029}\\
0.90 & $0.439 \pm 0.044$ & 0.403--0.496 & $0.533 \pm 0.052$ & 0.466--0.578 & 0.057\\
0.95 & $0.284 \pm 0.041$ & 0.248--0.339 & $\mathbf{0.397 \pm 0.048}$ & 0.347--0.461 & \textbf{0.029}\\
\bottomrule
\end{tabularx}
\end{adjustwidth}
\end{table}

Read this way the effect is larger than the AUC difference suggests. At a sensitivity of 0.80 the \emph{mix} policy raises specificity from 0.612 to 0.713, and at 0.95 from 0.284 to 0.397---gains of roughly 0.10 in both cases. At a fixed miss rate, in other words, the augmented model rejects about a quarter more of the benign lesions that the baseline would have referred. The 0.80 and 0.95 rows separate across seeds and reach the floor of the permutation test; the 0.90 row does not quite ($p = 0.057$), because one baseline run reaches 0.496 against a worst \emph{mix} run of 0.466. The direction is the same at all three targets.

\subsection{Where the Gain Comes From}

An aggregate AUC can hide a loss on a clinically important subgroup, so we broke the confirmation set down by \texttt{diagnosis\_3} at the transferred thresholds (Table~\ref{tab:perdx}). Most of the improvement comes from the non-malignant side: specificity rises on nevi (0.939 to 0.960) and on seborrheic keratosis (0.722 to 0.732), the categories the baseline most often mistook for malignancy. On the malignant side the changes are smaller and not uniform. Recall improves for invasive melanoma (0.391 to 0.424) and melanoma \emph{in situ} (0.261 to 0.282) but falls for the ``Melanoma, NOS'' group (0.450 to 0.419).

The low absolute recall in this table---between 0.26 and 0.45---should be read together with Section~\ref{sec:matched}. It reflects the transferred threshold rather than a ceiling on either model: both reach 0.95 sensitivity on the same data at a threshold chosen for that purpose, and the question is then what specificity they retain. What the per-diagnosis view does establish is where the two policies differ, and the answer is that the difference is concentrated on benign categories rather than distributed evenly across malignant subtypes.

\begin{table}[H]
\caption{Per-diagnosis breakdown on the confirmation set at the validation-tuned thresholds (baseline $0.476$, \emph{mix} $0.455$). Recall is reported for malignant subtypes and specificity for non-malignant ones; subtypes with fewer than 30 malignant or 100 non-malignant images are omitted. Labels are the raw \texttt{diagnosis\_3} values of the ISIC Archive.\label{tab:perdx}}
\begin{tabularx}{\textwidth}{Xlccc}
\toprule
\textbf{Diagnosis (\texttt{diagnosis\_3})} & \textbf{Class} & \textbf{\textit{n}} & \textbf{Baseline} & \textbf{Mix}\\
\midrule
Melanoma in situ & Malignant & 333 & 0.261 & \textbf{0.282}\\
Melanoma, NOS & Malignant & 329 & \textbf{0.450} & 0.419\\
Melanoma Invasive & Malignant & 238 & 0.391 & \textbf{0.424}\\
\midrule
Nevus & Non-malignant & 6628 & 0.939 & \textbf{0.960}\\
Seborrheic keratosis & Non-malignant & 489 & 0.722 & \textbf{0.732}\\
\bottomrule
\end{tabularx}
\end{table}

\subsection{External Clinical Cohort}\label{sec:external}

The Melanoscope collection tests transfer to an acquisition setting outside the ISIC Archive entirely. Across four runs per policy, ROC-AUC is $0.934 \pm 0.015$ for the baseline and $0.930 \pm 0.022$ for \emph{mix}: performance is maintained, and no difference is detectable. On the reference checkpoint pair the direction favours \emph{mix} in both AUC (0.916 to 0.941) and operating point (sensitivity 0.591 to 0.818 at specificity 0.980 to 0.967), but with 22 malignant cases that comparison does not reach significance ($p = 0.22$) and it is within the run-to-run spread of Table~\ref{tab:seeds}.

One descriptive observation is worth recording without being claimed. Run-to-run variability on this cohort is markedly lower under \emph{mix}: across the four runs per policy, AUPRC spans $0.743 \pm 0.040$ for the baseline and $0.775 \pm 0.010$ for \emph{mix}, and sensitivity at the tuned thresholds spans $0.659 \pm 0.176$ against $0.693 \pm 0.078$. With four runs per group a variance ratio of this size is not testable, so we report it as an observation rather than a finding; it would be worth checking directly with a larger number of runs.

The cohort is small. With 22 malignant cases it cannot resolve differences of the magnitude measured on the public sources, and this is the main limitation of the external evidence presented here. What it does establish is that an aggressive photometric and geometric policy---one that visibly distorts colour, contrast and local geometry during training---does not degrade performance when the model meets a genuinely new device and population. Larger external cohorts from several institutions are needed to move beyond that statement.

\subsection{What the Model Attends To}\label{sec:mechanism}

Two lines of evidence, both exploratory, are consistent with the model relying less on source-specific cues.

\begin{figure}[H]
\centering
\subfloat[\centering Lesion]{\includegraphics[width=0.31\textwidth]{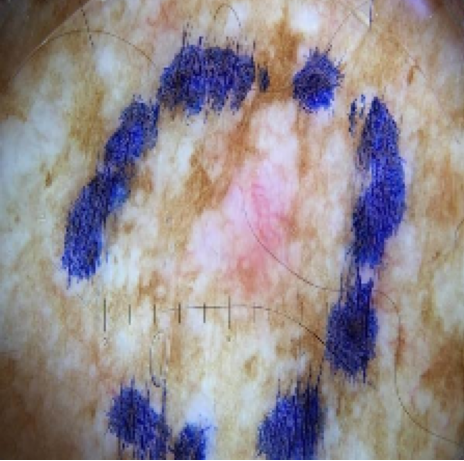}}
\hfill
\subfloat[\centering Baseline model]{\includegraphics[width=0.31\textwidth]{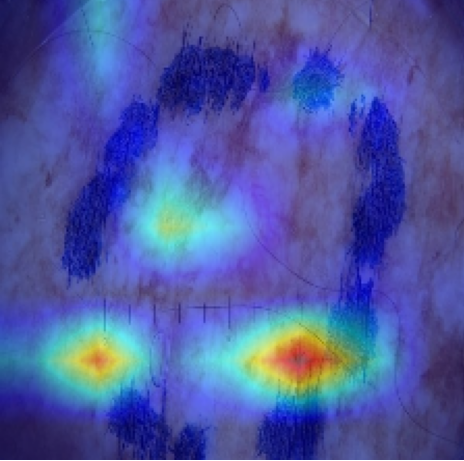}}
\hfill
\subfloat[\centering \emph{mix} policy]{\includegraphics[width=0.31\textwidth]{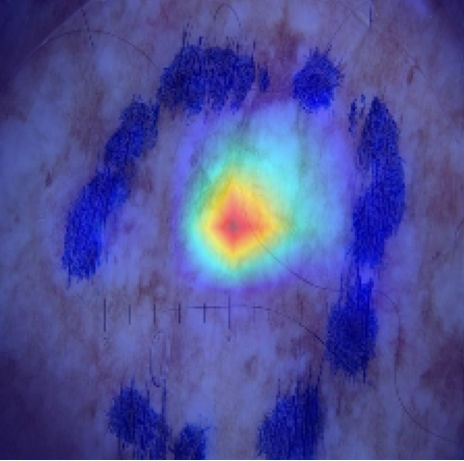}}
\caption{Grad-CAM comparison for a basal cell carcinoma case: (\textbf{a})~the input lesion; (\textbf{b})~baseline model, with attention spread over several diffuse, partly peripheral foci; (\textbf{c})~model trained with the selected augmentation policy, with a single compact activation centred on the lesion. Saliency is the gradient of the sigmoid logit with respect to the final convolutional feature map. This is a single illustrative case, not a quantitative result.\label{fig:gradcam}}
\end{figure}

Grad-CAM saliency maps were computed from the final convolutional stage on $224\times224$ inputs~\citep{ref-selvaraju}. For the illustrated basal cell carcinoma case (Figure~\ref{fig:gradcam}), the baseline spreads attention over several diffuse foci including peripheral regions away from the lesion, whereas the augmented model produces a single compact activation centred on it. This is one case, chosen for illustration; a panel-based reader study with agreement statistics would be required to treat it as evidence.

\begin{figure}[H]
\centering
\includegraphics[width=\textwidth]{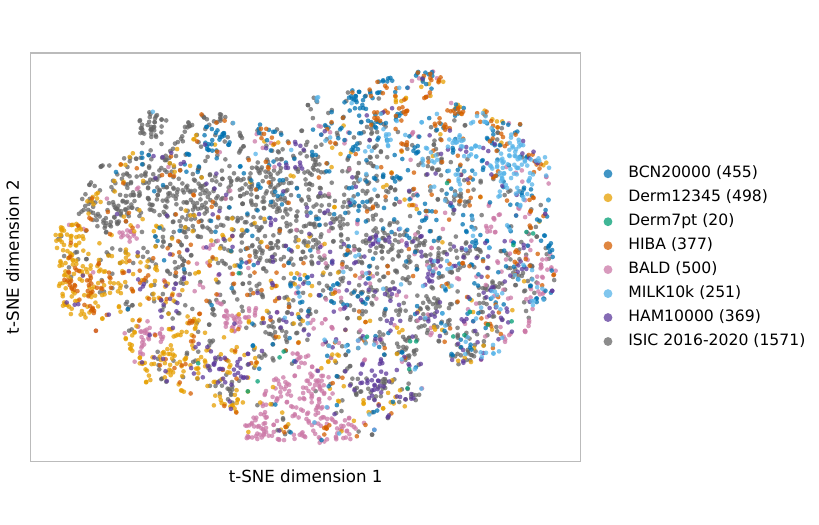}
\caption{t-SNE projection of test-set embeddings from ConvNeXt-Large trained with the selected augmentation policy, coloured by dataset, with the same palette and projection settings as Figure~\ref{fig:imagenet-tsne} so that the two are directly comparable. Points are more uniformly mixed across sources, though some grouping persists.\label{fig:tsne-after}}
\end{figure}

Embeddings from the \emph{mix}-trained model mix across sources more thoroughly than ImageNet-pretrained features, although residual grouping persists for BALD and Derm12345 (Figure~\ref{fig:tsne-after}). Quantifying this on the same embedding sets, the balanced accuracy of a 5-fold cross-validated logistic-regression source classifier falls from 0.819 to 0.764, and the silhouette score with respect to \texttt{attribution} falls from $-0.017$ to $-0.070$ (Table~\ref{tab:mixing}).

This comparison cannot attribute the reduction to augmentation. The two embedding sets differ in two respects at once: one is a general-purpose ImageNet representation, the other has been fine-tuned on the dermoscopic task \emph{with} augmentation. Fine-tuning alone reshapes a representation around lesion morphology and would reduce source separability by itself. The decisive comparison---a baseline fine-tuned without the search against \emph{mix}, on identical images at the same layer---requires embeddings from the baseline checkpoint that were not retained, and is left for future work.

\begin{table}[H]
\caption{Domain-mixing indices for ImageNet-pretrained and \emph{mix}-trained ConvNeXt-Large embeddings ($N=4041$, $D=1536$, 10 acquisition sources). Source-classifier balanced accuracy was estimated by 5-fold stratified cross-validation with logistic regression; the silhouette score uses \texttt{attribution} as the cluster label. Lower is better for both. \emph{Exploratory:} the two models differ both in fine-tuning and in augmentation, so the difference cannot be attributed to augmentation alone.\label{tab:mixing}}
\begin{tabularx}{\textwidth}{Xcc}
\toprule
\textbf{Model} & \textbf{Source-classifier balanced accuracy $\downarrow$} & \textbf{Silhouette score $\downarrow$}\\
\midrule
ImageNet pretrained & 0.819 & $-$0.017\\
\emph{Mix}-trained  & \textbf{0.764} & \textbf{$-$0.070}\\
\midrule
Chance level        & 0.100 & ---\\
\bottomrule
\end{tabularx}
\end{table}

\section{Discussion}

The augmentations that help most are those that model the actual physical cause of the shift. Photometric transformations dominate the useful set, and they do so because dermatoscopes, illumination, colour correction and white balance are what differ between the sources analysed in Section~\ref{sec:domainshift}. This is the expected shape of a domain-generalization result: the objective is not to fit one source well but to obtain a model that works on distributions different from the training one~\citep{ref-gulrajani}.

The size of the effect is worth stating plainly. On 8073 held-out images that took no part in choosing the policy and share no lesion and no institution with the training data, a change confined to the data pipeline buys $+0.039$ ROC-AUC, reproducibly across independent training runs and in the same direction on every contributing source. In-domain accuracy does not pay for it. Since the change is one block in a training script and requires neither source annotations nor target data, this is an unusually cheap intervention for its size.

The gain is not uniform across diagnoses, and the distribution matters clinically. Most of it comes from fewer false alarms on benign keratoses and nevi rather than from better melanoma detection, and recall on the largest melanoma subgroup decreases slightly. Held at a fixed miss rate, this translates into roughly a quarter fewer unnecessary referrals (Section~\ref{sec:matched}), which is a real benefit for a triage aid---but it is not the same benefit as catching more melanomas, and it should not be reported as though it were.

A gain on external data is not always accompanied by a gain in-domain, and that trade-off should be judged against the deployment scenario. For a system intended to scale to new institutions, a small in-domain gain matters less than robustness under external validation. Here the trade-off did not arise: in-domain performance was maintained.

These results also position augmentation relative to heavier machinery. Contrastive self-supervised pre-training can produce representations less sensitive to domain artifacts~\citep{ref-chen}; domain-adversarial training can penalize source-identifiable features when source labels are available~\citep{ref-ganin}; test-time adaptation can adjust to unlabeled target data at deployment~\citep{ref-sun,ref-wang-tent}. Each is more powerful in principle and each carries a requirement that augmentation does not. An augmentation policy is therefore the baseline such methods should be measured against, and we did not attempt a numerical comparison with them here.

\subsection{Limitations}\label{sec:limitations}

\textit{Selection remains source-level, not fully disjoint.} The confirmation set removes image-level selection bias: none of its 8073 images helped rank the policies. It does not remove source-level dependence, because those images come from ISIC~2016--2020, the same archives whose test splits formed the development set. A leave-one-source-out policy search, or selection on one archive with a single pre-registered comparison on another, would close the remaining gap and is the natural next experiment.

\textit{The external cohort is small.} Melanoscope is the only evaluation set outside the ISIC Archive, and with 472 images and 22 malignant cases it can establish that performance is maintained but cannot resolve a difference of the size measured on the public sources. Larger multi-institution cohorts are required.

\textit{The gain is unevenly distributed across diagnoses.} Most of the improvement is higher specificity on benign categories, and recall falls slightly for the largest melanoma subgroup. A deployed triage aid would be operated at a much higher sensitivity than the transferred threshold of Table~\ref{tab:confirm-op} produces; Section~\ref{sec:matched} reports the comparison at matched sensitivity, where the advantage is about 0.10 in specificity but the absolute specificity at 0.95 sensitivity is only 0.397. Neither model is close to a deployable operating point on these sources.

\textit{Methodological scope.} The statistical analysis covers the baseline and \emph{mix}; the other nine policies remain single-run point estimates, and no correction was applied for the eleven-way search. All experiments use one backbone and one data split, so architecture-level conclusions do not follow. BALD, Derm12345 and Derm7pt carry no lesion identifiers, so their splits are image-level and repeat imaging cannot be excluded within the in-domain partition; every held-out source does carry them. The interpretability evidence is exploratory, as stated in Section~\ref{sec:mechanism}. Finally, the binary formulation is useful for triage but conceals differences between melanoma, basal cell carcinoma, squamous cell carcinoma and premalignant conditions; a multi-class or hierarchical extension is a natural next step.

\section{Conclusions}

We searched systematically for data augmentations that make a dermoscopic malignancy classifier robust to domain shift, and measured the winner under a protocol that keeps the images used to choose a policy separate from the images used to judge it.

The selected \emph{mix} policy raised ROC-AUC from 0.787 to 0.826 on a confirmation set of 8073 held-out images that took no part in the selection and contain no lesion and no institution shared with the training data. The improvement held in every one of eight training runs, with non-overlapping per-seed ranges and an exact permutation $p$ at its attainable floor of 0.029, and it pointed the same way on each contributing source. Compared at a matched miss rate, which is how a triage aid is specified in practice, the same change raises specificity by about 0.10---roughly a quarter fewer unnecessary referrals at constant sensitivity. In-domain accuracy was preserved. On an independent clinical cohort acquired with a different device at a different institution, performance was maintained, though 22 malignant cases cannot resolve an effect of this size. Photometric transformations dominate the useful operations, which is consistent with colour and illumination being the measured axes of variation between sources.

Two directions follow. The first is a source-disjoint selection protocol---leave-one-source-out policy search, or selection on one archive with a pre-registered comparison on another---which would remove the last dependence between selection and evaluation. The second is external validation on larger multi-institution cohorts with a fixed annotation protocol, predefined quality thresholds and error analysis by lesion type, source and artifact. Beyond that, self-supervised pre-training and automated policy search are the natural extensions, and the results here give them a concrete baseline to beat.
\vspace{6pt}

\authorcontributions{Conceptualization, A.K. and Ev.K.; methodology, A.K. and I.L.; software, I.L. and E.U.; validation, I.L., El.K. and Ev.K.; formal analysis, I.L.; investigation, I.L. and El.K.; resources, O.S. and Ev.K.; data curation, El.K. and I.L.; writing---original draft preparation, I.L.; writing---review and editing, A.K., Ev.K. and O.S.; visualization, I.L.; supervision, A.K. and Ev.K.; project administration, A.K.; funding acquisition, A.K. All authors have read and agreed to the published version of the manuscript.}

\funding{This work was supported by a grant provided by the Ministry of Economic Development of the Russian Federation (agreement dated 20 June 2025, No. 139-15-2025-011, identifier 000000C313925P4G0002).}

%
\institutionalreview{This work is a retrospective, non-interventional analysis of pre-existing dermoscopic images; no prospective data collection from human participants or animals was carried out, and the study was conducted in accordance with the Declaration of Helsinki. Two categories of data were used. The ISIC Archive collections (BCN20000, Derm12345, HAM10000, ISIC~2016--2020, HIBA, BALD, MILK10k) and Derm7pt are public research datasets distributed in de-identified form under their respective licenses; ethical approval for the original collection of these images was obtained by the contributing institutions, and their secondary analysis required no further review. The Melanoscope collection is a non-public clinical dataset assembled and owned by the authors; it was de-identified before analysis and used exclusively to evaluate already-trained models---no image from it entered model training, hyperparameter selection, or threshold tuning. It was acquired within the clinical research programme ``Intelligent decision-support system for the diagnosis of skin neoplasms based on mobile dermatoscopy'', which was reviewed by the Local Ethics Committee of South-West State University, Kursk, Russia, and found to have been conducted in accordance with the principles of biomedical ethics of the Declaration of Helsinki (protocol No.~7 of 18 June 2025).}

\informedconsent{For the public ISIC Archive and Derm7pt collections, participant consent was obtained by the original contributing institutions as part of their own data-collection protocols; the present secondary analysis used only de-identified images and did not require additional consent. For the Melanoscope collection, written informed consent for the research use of the images was obtained from all patients at the time of acquisition.}

\dataavailability{This study uses a combination of public and closed clinical datasets. The ISIC Archive collections (BCN20000, Derm12345, HAM10000, ISIC~2016--2020, HIBA, BALD, MILK10k) are available through the International Skin Imaging Collaboration (\url{https://www.isic-archive.com}), and Derm7pt is available at \url{https://derm.cs.sfu.ca}. The external Melanoscope dataset is a closed clinical collection assembled and owned by the authors; it is not publicly available and cannot be redistributed, including on request, because the consent obtained from the participating patients covers research use by the collecting team rather than public release, and only aggregate performance figures derived from it are reported here. The training configuration files, the per-image prediction logits for the baseline and \emph{mix} models on every evaluation set and every training seed, and the single analysis script that rebuilds every number reported in this paper from those logits---including the pool definitions, the contamination filters, the bootstrap intervals, the DeLong tests and the exact permutation tests---are available from the corresponding author upon reasonable request.}

\acknowledgments{During the preparation of this manuscript, the authors used DeepL Translator (DeepL SE, web version, accessed July 2026) to translate portions of the text from Russian into English. The tool was applied only to the language of the manuscript; it was not used to generate, analyse, or interpret any scientific content, and it played no part in the design of the study, the experiments, or the statistical analysis. The authors reviewed and edited all translated text and take full responsibility for the content of this publication.}

\conflictsofinterest{The authors declare no financial conflicts of interest. As a potential non-financial competing interest, the authors note that they assembled and own the Melanoscope collection used for the external evaluation in Section~\ref{sec:external}, and are involved in the development of decision-support tools for dermoscopic diagnosis; the evaluated models, the augmentation policies, and the external dataset therefore originate from the same group. To limit the resulting risk of bias, the Melanoscope images were used only after model training and threshold selection had been completed and were never involved in any modelling decision, and all per-image predictions underlying the reported statistics have been retained. The funders had no role in the design of the study; in the collection, analyses, or interpretation of data; in the writing of the manuscript; or in the decision to publish the results.}

\abbreviations{Abbreviations}{
The following abbreviations are used in this manuscript:\\

\noindent
\begin{tabular}{@{}ll}
AI & Artificial Intelligence\\
AUPRC & Area Under the Precision--Recall Curve\\
CNN & Convolutional Neural Network\\
ISIC & International Skin Imaging Collaboration\\
OOD & Out-of-Domain\\
ROC-AUC & Area Under the Receiver Operating Characteristic Curve\\
t-SNE & t-Distributed Stochastic Neighbor Embedding\\
\end{tabular}
}

\appendixtitles{yes}
\appendixstart
\appendix

\section[\appendixname~\thesection]{}
\subsection{Composition of the Evaluation Levels}\label{app:dist}

Table~\ref{tab:dist-source} gives the per-source breakdown of the in-domain partition summarized in Table~\ref{tab:dist}, together with the two levels of the held-out evaluation defined in Section~\ref{sec:protocol}. Held-out counts are reported after the two contamination filters: 158 HAM10000 images sharing a lesion identifier with the MILK10k training or validation split, and 179 ISIC~2020 images attributed to the institution that contributed BCN20000, are excluded. HAM10000 contributes to the development level only, because its test split is the only part of that collection that entered the evaluation pool.

\begin{table}[H]
\caption{Per-source class distribution (non-malignant/malignant). ``B'' denotes non-malignant and ``M'' denotes malignant. Held-out sources are excluded from training entirely and contribute only to the development and confirmation levels.\label{tab:dist-source}}
\begin{tabularx}{\textwidth}{Xccc}
\toprule
\textbf{Source} & \textbf{Train (B/M)} & \textbf{Validation (B/M)} & \textbf{Test (B/M)}\\
\midrule
\multicolumn{4}{l}{\textit{In-domain sources}}\\
BCN20000 & 6066 / 6140 & 1748 / 1868 & 640 / 863\\
Derm12345 & 7721 / 788 & 2316 / 237 & 994 / 101\\
Derm7pt & 278 / 106 & 121 / 64 & 264 / 119\\
HIBA & 483 / 386 & 153 / 108 & 66 / 43\\
BALD & 365 / 101 & 110 / 30 & 47 / 13\\
MILK10k & 1059 / 2410 & 289 / 751 & 145 / 302\\
\midrule
\textbf{In-domain total} & \textbf{15{,}972 / 9931} & \textbf{4737 / 3058} & \textbf{2156 / 1441}\\
\midrule
\multicolumn{4}{l}{\textit{Held-out sources: development level}}\\
HAM10000 & --- & --- & 749 / 131\\
ISIC 2019 & --- & --- & 83 / 27\\
ISIC 2020 & --- & --- & 487 / 34\\
\textbf{Development total} & --- & --- & \textbf{1319 / 192}\\
\midrule
\multicolumn{4}{l}{\textit{Held-out sources: confirmation level}}\\
ISIC 2016 & --- & --- & 18 / 5\\
ISIC 2019 & --- & --- & 2264 / 525\\
ISIC 2020 & --- & --- & 4890 / 371\\
\textbf{Confirmation total} & --- & --- & \textbf{7172 / 901}\\
\bottomrule
\end{tabularx}
\end{table}

Table~\ref{tab:oodauc} breaks the confirmation-set result down by source for the reference checkpoint pair of Table~\ref{tab:confirm-auc}. Because both models see the same images, each row reports the \emph{paired} difference with a 95\% bootstrap interval rather than treating the two AUCs as independent. The corresponding multi-seed analysis is reported in Section~\ref{sec:seeds}.

\begin{table}[H]
\caption{Per-source breakdown of the confirmation set for the reference checkpoint pair ($N = 8073$). ``Prev.''\ is malignant prevalence. Each row gives the paired $\Delta$AUC with a 95\% bootstrap confidence interval (2000 resamples, resampling images jointly for the two models). Rows whose interval excludes zero are shown in bold; no correction is applied for testing three sources.\label{tab:oodauc}}
\begin{adjustwidth}{-\extralength}{0cm}
\small
\begin{tabularx}{\fulllength}{Xccccccc}
\toprule
\textbf{Source} & \textbf{\textit{N}} & \textbf{\textit{N}\textsubscript{mal}} & \textbf{Prev.} & \textbf{Baseline AUC} & \textbf{Mix AUC} & \textbf{$\Delta$AUC (95\% CI)}\\
\midrule
ISIC 2016 &   23 &    5 & 0.217 & 0.856 & 0.900 & $+$0.044 ($-$0.089--$+$0.211)\\
ISIC 2019 & 2789 &  525 & 0.188 & 0.738 & 0.766 & \textbf{$+$0.028 ($+$0.012--$+$0.042)}\\
ISIC 2020 & 5261 &  371 & 0.070 & 0.792 & 0.849 & \textbf{$+$0.057 ($+$0.043--$+$0.072)}\\
\midrule
\textbf{Total} & \textbf{8073} & \textbf{901} & \textbf{0.112} & 0.786 & 0.832 & \textbf{$+$0.047 ($+$0.037--$+$0.056)}\\
\bottomrule
\end{tabularx}
\end{adjustwidth}
\end{table}

\subsection{Sensitivity to the Contamination Filters}\label{app:filters}

The two filters of Section~\ref{sec:protocol} remove 337 of the 9921 images in the raw held-out pool. Table~\ref{tab:filters} reports the comparison with and without them, on the pooled held-out data, so that the effect of the filtering can be read directly. Removing the contaminated images raises both models and leaves the difference between them essentially unchanged; the filters therefore make the evaluation cleaner without manufacturing the effect. The 158 lesion-overlap images are, if anything, easier for the baseline than the rest of HAM10000, which is consistent with the overlap diluting the domain shift rather than favouring either policy.

\begin{table}[H]
\caption{Effect of the contamination filters on the pooled held-out evaluation. ``Raw pool'' is the unfiltered set of held-out images; ``filtered pool'' removes the 158 HAM10000 images sharing a lesion identifier with MILK10k training or validation data and the 179 ISIC~2020 images attributed to the institution that contributed BCN20000. Multi-seed figures are means over four runs per policy; reference figures are the retained checkpoint pair with 95\% bootstrap intervals on the difference.\label{tab:filters}}
\begin{adjustwidth}{-\extralength}{0cm}
\small
\begin{tabularx}{\fulllength}{Xcccccc}
\toprule
& & \multicolumn{3}{c}{\textbf{Multi-seed (mean over 4 runs)}} & \multicolumn{2}{c}{\textbf{Reference pair}}\\
\cmidrule(lr){3-5}\cmidrule(lr){6-7}
\textbf{Pool} & \textbf{\textit{N}} & \textbf{Baseline} & \textbf{Mix} & \textbf{$\Delta$} & \textbf{$\Delta$AUC} & \textbf{95\% CI}\\
\midrule
Raw pool      & 9921 & 0.770 & 0.817 & $+$0.047 & $+$0.053 & $+$0.045--$+$0.061\\
Filtered pool & 9584 & 0.792 & 0.836 & $+$0.044 & $+$0.052 & $+$0.044--$+$0.061\\
\bottomrule
\end{tabularx}
\end{adjustwidth}
\end{table}

\section[\appendixname~\thesection]{}
\subsection{Augmentation Configurations}\label{app:aug}

\begin{table}[H]
\caption{Color-augmentation combinations tested in the search.\label{tab:color}}
\begin{tabularx}{\textwidth}{lX}
\toprule
\textbf{Set} & \textbf{Composition}\\
\midrule
Color 1 & color\_jitter + hestain\\
Color 2 & color\_jitter + hestain + hsv\\
Color 3 & color\_jitter + hsv\\
Color 4 & planckian\_jitter + color\_jitter\\
Color 5 & planckian\_jitter + color\_jitter + hsv\\
Color 6 & planckian\_jitter + color\_jitter + hsv + hestain\\
Color 7 & planckian\_jitter + hestain\\
Color 8 & planckian\_jitter + hestain + color\_jitter\\
Color 9 & planckian\_jitter + hestain + hsv\\
Color 10 & planckian\_jitter + hsv\\
Color 11 & hestain + hsv\\
\bottomrule
\end{tabularx}
\end{table}

{\footnotesize
\begin{longtable}{@{}l p{9.8cm}@{}}
\caption{Composition of the final augmentation configurations.\label{tab:configs}}\\
\toprule
\textbf{Configuration} & \textbf{Included Augmentations}\\
\midrule
\endfirsthead
\multicolumn{2}{l}{\footnotesize\textit{Table \thetable{} continued.}}\\
\toprule
\textbf{Configuration} & \textbf{Included Augmentations}\\
\midrule
\endhead
\midrule
\multicolumn{2}{r}{\footnotesize\textit{continued on next page}}\\
\endfoot
\bottomrule
\endlastfoot
simple & Transpose, VerticalFlip, HorizontalFlip, GridDistortion, OpticalDistortion, ColorJitter, PlanckianJitter, HueSaturationValue, ToGray, ChromaticAberration, RandomGamma, CLAHE\\
\midrule
prime & Transpose, VerticalFlip, HorizontalFlip, Rotate, MotionBlur, MedianBlur, GaussianBlur, GaussNoise, GridDistortion, OpticalDistortion, ColorJitter, PlanckianJitter, HueSaturationValue, ToGray, ChromaticAberration, RandomGamma, CLAHE\\
\midrule
strong\_hestain & Transpose, VerticalFlip, HorizontalFlip, GridDistortion, OpticalDistortion, ColorJitter, PlanckianJitter, HueSaturationValue, ToGray, ChromaticAberration, RandomGamma, CLAHE, HEStain\\
\midrule
kaggle & Transpose, VerticalFlip, HorizontalFlip, Rotate, MotionBlur, MedianBlur, GaussianBlur, GaussNoise, GridDistortion, OpticalDistortion, ShiftScaleRotate, ElasticTransform, ColorJitter, PlanckianJitter, HueSaturationValue, ToGray, ChromaticAberration, RandomGamma, CLAHE\\
\midrule
kaggle\_simple & Transpose, VerticalFlip, HorizontalFlip, Rotate, MotionBlur, MedianBlur, GaussianBlur, GaussNoise, GridDistortion, OpticalDistortion, ColorJitter, PlanckianJitter, HueSaturationValue, ToGray, ChromaticAberration, RandomGamma, CLAHE\\
\midrule
kaggle\_strong & Transpose, VerticalFlip, HorizontalFlip, Rotate, MotionBlur, MedianBlur, GaussianBlur, GaussNoise, GridDistortion, OpticalDistortion, ShiftScaleRotate, ElasticTransform, Affine, ColorJitter, PlanckianJitter, HueSaturationValue, ToGray, ChromaticAberration, RandomGamma, CLAHE\\
\midrule
kaggle\_strong\_hestain & Transpose, VerticalFlip, HorizontalFlip, Rotate, MotionBlur, MedianBlur, GaussianBlur, GaussNoise, GridDistortion, OpticalDistortion, ShiftScaleRotate, ElasticTransform, Affine, ColorJitter, PlanckianJitter, HueSaturationValue, ToGray, ChromaticAberration, RandomGamma, CLAHE, HEStain\\
\midrule
mix & Transpose, VerticalFlip, HorizontalFlip, Affine, Rotate, CLAHE, MotionBlur, MedianBlur, GaussianBlur, GaussNoise, OpticalDistortion, GridDistortion, ElasticTransform, ColorJitter, PlanckianJitter, HueSaturationValue, HEStain, ToGray, ChromaticAberration, RandomGamma\\
\midrule
combine\_all & Transpose, VerticalFlip, HorizontalFlip, Rotate, MotionBlur, MedianBlur, GaussianBlur, GaussNoise, GridDistortion, OpticalDistortion, ShiftScaleRotate, GridDropout, ElasticTransform, ColorJitter, PlanckianJitter, RandomBrightnessContrast, HueSaturationValue, ToGray, ChromaticAberration, RandomGamma, CLAHE, HEStain\\
\midrule
low\_intensity\_prime & Transpose, VerticalFlip, HorizontalFlip, Rotate, MotionBlur, MedianBlur, GaussianBlur, GaussNoise, GridDistortion, OpticalDistortion, ColorJitter, PlanckianJitter, HueSaturationValue, ToGray, ChromaticAberration, RandomGamma, CLAHE\\
\midrule
low\_intensity\_kaggle\_strong & Transpose, VerticalFlip, HorizontalFlip, Rotate, MotionBlur, MedianBlur, GaussianBlur, GaussNoise, GridDistortion, OpticalDistortion, ShiftScaleRotate, ElasticTransform, Affine, ColorJitter, PlanckianJitter, RandomBrightnessContrast, HueSaturationValue, ToGray, ChromaticAberration, RandomGamma, CLAHE\\
\end{longtable}
}

Table~\ref{tab:configs} lists only which operators each policy contains. Several policies share an identical operator set and are distinguished solely by application probabilities, parameter ranges, and the grouping of operators into mutually exclusive \texttt{OneOf} blocks: this is the case for \emph{prime}, \emph{kaggle\_simple}, and \emph{low\_intensity\_prime}, and likewise for \emph{kaggle} and \emph{mix}. The \emph{low\_intensity\_*} variants use the same operators as their parent policy with the magnitude ranges narrowed. Because these details determine the results in Table~\ref{tab:results}, the complete specification of the selected \emph{mix} policy is given in Table~\ref{tab:mixspec}; the corresponding specifications for the remaining policies are available from the corresponding author on request. All augmentations were implemented with \texttt{albumentations} version~2.0.8, applied in the listed order before resizing and normalization, and sampled independently per image with the seed of the training run. The version matters for exact reproduction, because several operator signatures used here (for example the \texttt{std\_range} parameterization of \texttt{GaussNoise} and the \texttt{fill}/\texttt{keep\_ratio} arguments of \texttt{Affine}) belong to the 2.x API and differ from their 1.x equivalents.

{\footnotesize
\begin{longtable}{@{}l c p{6.4cm}@{}}
\caption{Complete specification of the selected \emph{mix} policy, in application order. ``$p$'' is the probability of applying the operator. \texttt{OneOf} blocks select exactly one of their members when the block fires; the per-member probabilities inside a block are relative weights.\label{tab:mixspec}}\\
\toprule
\textbf{Operator} & \textbf{\textit{p}} & \textbf{Parameters}\\
\midrule
\endfirsthead
\multicolumn{3}{l}{\footnotesize\textit{Table \thetable{} continued.}}\\
\toprule
\textbf{Operator} & \textbf{\textit{p}} & \textbf{Parameters}\\
\midrule
\endhead
\midrule
\multicolumn{3}{r}{\footnotesize\textit{continued on next page}}\\
\endfoot
\bottomrule
\endlastfoot
Transpose & 0.5 & ---\\
VerticalFlip & 0.5 & ---\\
HorizontalFlip & 0.5 & ---\\
Affine & 0.4 & translate $\pm10\%$; scale $0.85$--$1.15$; rotate $\pm360^\circ$; shear $0$; border constant, fill $0$\\
Rotate & 0.8 & limit $\pm360^\circ$, \texttt{crop\_border} = true\\
CLAHE & 0.4 & \texttt{clip\_limit} $=4$, tile grid $8\times8$\\
\midrule
\multicolumn{3}{@{}l}{\textit{OneOf} (blur/noise), $p = 0.7$:}\\
\quad MotionBlur & 1.0 & \texttt{blur\_limit} $=5$\\
\quad MedianBlur & 1.0 & \texttt{blur\_limit} $=5$\\
\quad GaussianBlur & 1.0 & \texttt{blur\_limit} $=5$\\
\quad GaussNoise & 1.0 & \texttt{std\_range} $=0.02$--$0.11$\\
\midrule
\multicolumn{3}{@{}l}{\textit{OneOf} (geometric distortion), $p = 0.7$:}\\
\quad OpticalDistortion & 1.0 & \texttt{distort\_limit} $=\pm1.0$\\
\quad GridDistortion & 1.0 & \texttt{num\_steps} $=5$, \texttt{distort\_limit} $=\pm1.0$\\
\quad ElasticTransform & 1.0 & $\alpha = 3$\\
\midrule
\multicolumn{3}{@{}l}{\textit{OneOf} (photometric), $p = 0.7$; each of the first five members is a two-operator \texttt{Compose}:}\\
\quad ColorJitter + PlanckianJitter & 0.5 / 0.3 & see below\\
\quad ColorJitter + HueSaturationValue & 0.5 / 0.3 & see below\\
\quad HEStain + PlanckianJitter & 0.2 / 0.3 & see below\\
\quad HEStain + ColorJitter & 0.2 / 0.5 & see below\\
\quad HEStain + HueSaturationValue & 0.2 / 0.3 & see below\\
\quad PlanckianJitter & 0.3 & see below\\
\quad ColorJitter & 0.5 & see below\\
\quad ToGray & 0.001 & ---\\
\midrule
\multicolumn{3}{@{}l}{\textit{Shared photometric parameters:}}\\
\quad ColorJitter & --- & brightness, contrast, saturation $0.9$--$1.1$; hue $\pm0.1$\\
\quad PlanckianJitter & --- & \texttt{temperature\_limit} $=3000$--$11\,000$~K\\
\quad HueSaturationValue & --- & hue shift $\pm5$; saturation shift $\pm30$; value shift $\pm30$\\
\quad HEStain & --- & \texttt{method} = preset, \texttt{preset} = \texttt{macenko}; intensity scale $0.9$--$1.1$, intensity shift $\pm0.1$\\
\midrule
ChromaticAberration & 0.5 & primary and secondary distortion limits $\pm0.1$, mode random\\
RandomGamma & 0.5 & \texttt{gamma\_limit} $=40$--$160$\\
\midrule
Resize & 1.0 & $224\times224$\\
Normalize & 1.0 & mean $(0.658, 0.542, 0.499)$, std $(0.237, 0.217, 0.219)$\\
\end{longtable}
}

\begin{adjustwidth}{-\extralength}{0cm}

\reftitle{References}

\isAPAandChicago{}{%

}

\end{adjustwidth}
\end{document}